\documentclass[nohyperref]{article}

\usepackage{microtype}
\usepackage{graphicx}
\usepackage{subfigure}
\usepackage{booktabs} % for professional tables

\usepackage{hyperref}

\usepackage[accepted]{icml2022}

\usepackage{amsmath}
\usepackage{amssymb}
\usepackage{mathtools}
\usepackage{amsthm}
\usepackage{optidef}
\usepackage{thmtools}
\usepackage{thm-restate}

\usepackage[capitalize,noabbrev]{cleveref}

\theoremstyle{plain}
\newtheorem{theorem}{Theorem}[section]

\theoremstyle{definition}

\theoremstyle{remark}
\newtheorem{remark}[theorem]{Remark}

\newcommand{\eg}{\textit{e}.\textit{g}. }
\newcommand{\ie}{\textit{i}.\textit{e}. }
\newcommand{\opt}{\hat{\theta}(\mathbf{1})}
\newcommand{\hess}{\mathbf{H}_{\opt}^{-1}}

\DeclareMathOperator*{\argmin}{arg\,min}

\newcommand{\RR}{\mathbb{R}}

\newcommand{\Exp}{\mathbb{E}}

\usepackage[textsize=tiny]{todonotes}

\icmltitlerunning{Achieving Fairness at No Utility Cost via Data Reweighing with Influence}

\begin{document}

\twocolumn[
\icmltitle{Achieving Fairness at No Utility Cost via Data Reweighing with Influence}



\icmlsetsymbol{equal}{*}

\begin{icmlauthorlist}
\icmlauthor{Peizhao Li}{yyy}
\icmlauthor{Hongfu Liu}{yyy}
% \icmlauthor{Firstname3 Lastname3}{comp}
\end{icmlauthorlist}

\icmlaffiliation{yyy}{Brandeis University}

\icmlcorrespondingauthor{Peizhao Li}{peizhaoli@brandeis.edu}

\icmlkeywords{Machine Learning, ICML}

\vskip 0.3in
]



\printAffiliationsAndNotice{}  % leave blank if no need to mention equal contribution

\begin{abstract}
With the fast development of algorithmic governance, fairness has become a compulsory property for machine learning models to suppress unintentional discrimination. In this paper, we focus on the pre-processing aspect for achieving fairness, and propose a data reweighing approach that only adjusts the weight for samples in the training phase. Different from most previous reweighing methods which usually assign a uniform weight for each (sub)group, we granularly model the influence of each training sample with regard to fairness-related quantity and predictive utility, and compute individual weights based on influence under the constraints from both fairness and utility. Experimental results reveal that previous methods achieve fairness at a non-negligible cost of utility, while as a significant advantage, our approach can empirically release the tradeoff and obtain cost-free fairness for equal opportunity. We demonstrate the cost-free fairness through vanilla classifiers and standard training processes, compared to baseline methods on multiple real-world tabular datasets. Code available at \href{https://github.com/brandeis-machine-learning/influence-fairness}{https://github.com/brandeis-machine-learning/influence-fairness}.
\end{abstract}

%-------------------------------------------------------------------------

\section{Introduction}
\label{sec:intro}

For artificial intelligence technology deployed in high-stakes applications like welfare distribution or school admission, it is essential to regulate algorithms and prevent unaware discrimination and unfairness in decision making~\cite{barocas2016big,goodman2017european,ferguson2017rise,civil}. Even though general data-driven algorithms are not designed to be unfair, the outcomes can still violate the AI principle of equality unintentionally~\cite{chouldechova2017fair}. Typically learning from historically biased data, the learner can retain or even amplify the inherent bias if there is no proper constraint on data or algorithms. As a consequence, the decisions from these algorithms may disadvantage users in certain sensitive groups (\eg female and African Americans), therefore raising societal concerns.

To mitigate unfairness algorithmically, solutions can be divided into three main categories: pre-processing, in-processing, and post-processing. Pre-processing approaches adjust the input data or training sample weights, and expect a vanilla learner can deliver fair results from the fair data transformation~\cite{kamiran2012data,krasanakis2018adaptive,calmon2017optimized,jiang2020identifying,feldman2015certifying,yan2020fair,zemel2013learning,rastegarpanah2019fighting,chhabra2021fair}. In-processing approaches insert fair constraints or penalties into the training pipeline, so the fair performance can be generalized to inference as achieved during training~\cite{zhang2018mitigating,agarwal2018reductions,zhao2019conditional,zafar2017fairness,jiang2020wasserstein,kearns2018preventing,goh2016satisfying,Li_2020_CVPR,li2021on,10.1145/3447548.3467225}. As a drawback, adding non-convex constraints or balancing the additional fair penalties with the primary training objective could rise optimization complexity, and sometimes incur instability~\cite{cotter2019optimization,roth2017stabilizing}. Post-processing approaches calibrate the outcomes independently from a model~\cite{hardt2016equality,pleiss2017fairness}, but would result in sub-optimal solutions~\cite{woodworth2017learning} and request sensitive attributes during the inference stage.

In this work, we advocate the pre-processing category since it directly diagnoses and corrects the source of bias, and can be easily adapted to existing data analytic pipelines. Pre-processing approaches can be further divided into two subcategories. The first one modifies original data or injects extra antidote data~\cite{calmon2017optimized,feldman2015certifying}, but will expose to the risk of learning from falsification which could be legally questionable~\cite{barocas2016big,krasanakis2018adaptive}. Conservatively, we consider sample reweighing, the second subcategory that purely adjusts the weights for samples in the training objective, and intend to achieve fairness through vanilla classifiers.

To prevent the model from disadvantaging a certain sensitive group and violating group-level fairness~\cite{dwork2012fairness,hardt2016equality}, most previous reweighing methods statistically identify the unprivileged groups from a heuristic or a learning process. Then, they assign an identical weight to all samples in a sensitive (sub)group, where the group membership is conditional on both sensitive attributes and the target label~\cite{kamiran2012data,jiang2020identifying}. The philosophy is to amplify the error from an underrepresented group in training, so optimization can equally update a model for different groups. Differently, we hypothesize the inherent bias in data can be traced to a biased and unconfident labeling function~\cite{chen2018my,jiang2020identifying}. The labeling might be unconsciously affected by sensitive attributes and therefore lead some training samples to be assigned with an improper label. By this means, some unqualified samples are assigned with positive labels, camouflaging themselves to be privileged, and correspondingly, some qualified samples are labeled as negative, inducing their groups to be unprivileged. Instead of equally treating every sample, we believe that an ideal pre-processing method is to downweight the mislabeled samples in training while keeping other good samples unchanged to preserve the predictive utility.

To this end, we propose a one-pass data reweighing method that granularly computes a weight for every sample in the training set. We use influence function~\cite{cook1980characterizations,hampel2011robust,koh2017understanding,giordano2019swiss} from robust statistics to estimate the effect of changing the weight of one sample without explicitly retraining the model. Specifically, we measure the sample influence in terms of both fairness and predictive utility, and theoretically prove that in a general case on influential approximation, a soft downweighting over some training samples can always enhance fairness while avoiding a cost in utility. We realize our findings through a proposed reweighing algorithm that estimates the individual weight through linear programming under both utility and fairness constraints. On multiple tabular datasets and in most cases, our empirical results achieve good group fairness at no utility cost compared to vanilla classifiers with original training data. We deem fairness at no utility cost as a significant advantage since it could help to popularize fair algorithms for extensive utility-driven products.

% Previous empirical results reveal that algorithmic fairness is generally achieved with a negligible sacrifice on predictive utility~\cite{agarwal2018reductions,chouldechova2017fair,feldman2015certifying}. This fact could bring concerns for fair algorithms because fairness is not always the first priority to pursue for a utility-driven product. As a significant advantage, our reweighing approach releases the tension between fairness and utility, and achieves a good fairness over multiple tabular datasets with no utility cost.

% We find that previous data reweighing methods for algorithmic fairness~\cite{jiang2020identifying,kamiran2012data,krasanakis2018adaptive} tend to reweigh instances from the same group (sensitive groups or groups formed by ground-truth label) with a same weight. We hypothesize that, in a binary classification case, the unfairness in data can be traced to the unfair labeling function, where qualified instances in unprivileged group are partially labeled as negative, and likewise, unqualified instances in privileged group are partially assigned with a positive outcome. Though the qualification here is ambiguous, an ideal pre-processing technique is expected to weigh down the unfairly labeled instances in the training objective, both in unprivileged and privileged groups. An unified weighing strategy at group-level prevents the pre-processing granularly weighing the training instances.

%-------------------------------------------------------------------------

\section{Characterizing Sample Influence}
\label{sec:influence}

In this section, we introduce preliminaries on influence function~\cite{cook1980characterizations,hampel2011robust,koh2017understanding,giordano2019swiss}. Influence function from robust statistics is to quantitatively measure the impact of an infinitesimal fraction of samples to an estimator. Consider a classifier with parameters $\theta\in\RR^{D}$ mapping instances from input space $x\in\mathcal{X}$ to output space $y\in\mathcal{Y}$. The model is trained on a training set $\mathcal{T} = \{z_i=(x_i, y_i)\}_{i=1}^{N_\mathcal{T}}$ with some loss function $\ell:\mathcal{X}\times\mathcal{Y}\times\Theta\rightarrow\RR$ through empirical risk minimization on the training set:
\begin{equation}\label{eq:train}
    \hat{\theta}(\mathbf{1}) = \argmin_{\theta}\sum_{i} \ell(z_i;\theta).
\end{equation}
The all-one vector $\mathbf{1}$ indicates an equal assignment of a unit weight to every training sample. A reweighing of samples followed by a retraining can be expressed as
\begin{equation}
\label{eq:retrain}
    \hat{\theta}(\mathbf{1} - \mathbf{w}) = \argmin_{\theta}\sum_{i} (1 - w_i) \ell(z_i;\theta),
\end{equation}
where $w_i$ denotes the deviation from a unit weight for $z_i$, and $\mathbf{w}\in\RR^{N_\mathcal{T}}$. Note that here $(1-w_i)$ is the weight of $z_i$, rather than $w_i$. A larger $w_i$ indicates its less importance of $z_i$ to the model training, and $w_i=1$ is equivalents to an entire removal of $z_i$ from the training set.

It is of high interest and value to know in a counterfactual how the model will change with regard to some typical measurements, \eg fairness or utility, if there is a reweighing. The actual influence derived from a reweighing can be expressed as follows:
\begin{equation}
    \mathcal{I}^*_f(\mathbf{w}) = f(\hat{\theta}({\mathbf{1}} - \mathbf{w})) - f(\hat{\theta}(\mathbf{1})),
\end{equation}
where the function $f:\RR^{D}\rightarrow\RR$ evaluates the quantity of interest. In the next section, we shall realize $f$ with functions describing utility and fairness, and study the correlations between these two influences. To compute $\mathcal{I}^*$, one can always separately train the model two times with and without the reweighing. However, to reach an ideal value in $f$, retraining is a brute force approach and can be prohibitively expensive to find the optimal $\mathbf{w}$. To this end, a first-order approximation called influence function helps to estimate the actual influence $\mathcal{I}^*$ while getting rid of retraining~\cite{cook1980characterizations,hampel2011robust}.

Influence function measures the effect of changing an infinitesimal weight from samples in $\mathbf{w}$, then linearly extrapolates to complete all of $\mathbf{w}$. It assumes $\ell$ to be twice-differentiable and strictly convex in $\theta$, and $f$ to be differentiable as well. These assumptions are mild and feasible for many pipelines involving classifiers like logistic regression to process tabular data. Having $\theta(\mathbf{1})$ and $\theta(\mathbf{1}-\mathbf{w})$ satisfied their first-order optimality conditions, and by taking a Taylor approximation, the actual influence $\mathcal{I}^*_f(\mathbf{w})$ can be approximated by an estimation
\begin{equation}
\label{eq:infl}
\begin{aligned}
    \mathcal{I}&_f(\mathbf{w}) = \nabla_\theta f(\hat{\theta}(\mathbf{1}))^\top\left[\frac{d}{dt}\theta(\mathbf{1}-t\mathbf{w})|_{t=0}\right] \\ &= \nabla_\theta f(\hat{\theta}(\mathbf{1}))^\top \mathbf{H}_{\hat{\theta}(\mathbf{1})}^{-1}\left[\sum_{i}w_i\nabla_\theta\ell(z_i;\hat{\theta}(\mathbf{1}))\right],
\end{aligned}
\end{equation}
where $\mathbf{H}_{\hat{\theta}(\mathbf{1})}=\sum_{i=1}^{N_\mathcal{T}}\nabla_\theta^2\ell(z_i;\hat{\theta}(\mathbf{1}))$ is the Hessian matrix of $\ell$, and the convexity ensures its invertibility. Influence function has been empirically proofed to be valid for representing the actual influence obtained by model retraining when the above assumptions are satisfied or violated by a small degree~\cite{koh2017understanding}. The additivity of $\mathcal{I}_f(\mathbf{w})$ regarding a set of $\mathbf{w}$ should follow the additivety of $f$. A negative value of $\mathcal{I}_f(\mathbf{w})$ instructs a reduction in $f$ if the model  $\opt$ is retrained with the weight $\mathbf{w}$ by~\cref{eq:retrain}.

%-------------------------------------------------------------------------

\section{Fairness at No Utility Cost}

We consider group fairness in this work. Group fairness articulates the equality of some statistics like predictive rate or true positive rate between certain groups. And here a group is specifically constructed based on some sensitive attributes like gender or race. Consider a binary classification problem $\mathcal{Y}=\{0,1\}$, where predictions are affected by a binary sensitive attribute $a\in\{0,1\}$. The sensitive attribute divides samples into a privileged group and an unprivileged group. An initial notion of group fairness is called Demographic Parity~\cite{dwork2012fairness}, requesting equality on the rate of positive predictions:
\begin{equation}
    \Pr(\hat{y}\ |\ a=0)=\Pr(\hat{y}\ |\ a=1).
\end{equation}
Demographic Parity enforces a group fairness on the outcomes regardless of the gap in base rate, \ie $\Pr(y\ |\ a=0)\neq\Pr(y\ |\ a=1)$. If sensitive groups have a gap in the base rate, there will always be a tradeoff between DP and utility for any classifier. To this end, another notion called Equal Opportunity~\cite{hardt2016equality} has been raised to measure the equality on true positive rate:
\begin{equation}
    \Pr(\hat{y}\ |\ a=0,y=1)=\Pr(\hat{y}\ |\ a=1,y=1).
\end{equation}
Other notions~\cite{mehrabi2021survey} also help to characterize the fairness problem including Equalized Odds (equality on both true positive rate and true negative rate), Accuracy Parity (equality on the predictive error rate), Predictive Equality (equality on the false positive rate), etc. We focus on Equal Opportunity and Demographic Parity while the rest notions could be incremental extensions to our framework.

To formulate the fairness issue as an optimization problem, one can intuitively quantify the inequality and turn the subjection of fairness into an objective function. For instance, with $\ell_{0/1}$ denotes zero-one loss, the gap in Equal Opportunity is:
\begin{equation}
    \Big|\Exp\left[\ell_{0/1}(z;\theta)|a=1,y=1\right] - \Exp\left[\ell_{0/1}(z;\theta)|a=0,y=1\right]\Big|.
\end{equation}
To make the function differentiable, a surrogate function is necessary to replace $\ell_{0/1}$ and here we substitute it with the training loss $\ell$ for Equal Opportunity, following previous works~\cite{zafar2017fairness,donini2018empirical}. Demographic Parity can be derived similarly. The fair loss over a sample set $\mathcal{S}$ for these two notions are:
\begin{equation}
\label{eq:fair}
\begin{aligned}
    f_\text{eop}^{\mathcal{S}}(\theta)&=\Big|\Exp_\mathcal{S}\left[\ell(z;\theta)\ |\ a=1,y=1\right] \\&-\Exp_\mathcal{S}\left[\ell(z;\theta)\ |\ a=0,y=1\right]\Big|, \\
    f_\text{dp}^{\mathcal{S}}(\theta)&=\Big|\Exp_\mathcal{S}\left[\hat{y}\ |\ a=1\right] - \Exp_\mathcal{S}\left[\hat{y}\ |\ a=0\right]\Big|.
\end{aligned}
\end{equation}
We now begin to elaborate on our framework. Instead of evaluating the fairness and utility on the training set $\mathcal{T}$, we are interested in the change of this two-side performance for the classifier on a validation set $\mathcal{V}=\{z_j=(x_j,y_j)\}_{j=1}^{N_\mathbf{v}}$ before and after a reweighing over $\mathcal{T}$, where each sample $z_j$ in the validation set is associated with a binary sensitive attribute $a_j\in\{0,1\}$. Measuring the influence of training samples on $\mathcal{V}$ towards fairness and utility can help us identify which sample has a positive impact on training, and to be precise, in which direction and to what extent.

To clarify the source of a sample, we use $i$ to denote the index belonging to the training set, and $j$ for the validation set. Let $\mathbf{e}_i$ stand for an all-zero vector except entry $i$ equal to 1. Deploying $\mathbf{e}_i$ as a weight for retraining means a hard removal of $z_i$ while preserving the rest samples unchanged. The influence function on fairness can be derived by realizing the function $f$ in~\cref{eq:infl} with~\cref{eq:fair}:
\begin{equation}
\label{eq:infl_fair}
    \mathcal{I}_\text{eop/dp}(\mathbf{e}_i) = {\nabla_{\theta}f_\text{eop/dp}^\mathcal{V}(\hat{\theta}(\mathbf{1}))}^\top\mathbf{H}_{\hat{\theta}(\mathbf{1})}^{-1}\nabla_{\theta}\ell(z_i;\hat{\theta}(\mathbf{1})),
\end{equation}
and similarly, the influence on utility is expressed as
\begin{equation}
\label{eq:infl_util}
    \mathcal{I}_\text{util}(\mathbf{e}_i) = \sum_{j}\nabla_{\theta}\ell(z_j;\hat{\theta}(\mathbf{1}))^\top\mathbf{H}_{\hat{\theta}(\mathbf{1})}^{-1}\nabla_{\theta}\ell(z_i;\hat{\theta}(\mathbf{1})).
\end{equation}
Here we use a truncated subscript `eop/dp' to express the feasibility for either functions, and in what follows we may use $f_\text{fair}^\mathcal{V}$ to unify $f_\text{eop/dp}^\mathcal{V}$. The influence function satisfies additivity followed by the additivity of $\nabla_{\theta}f_\text{fair}^\mathcal{V}$ and $\sum_{j}\nabla_{\theta}\ell(z_j;\hat{\theta(\mathbf{1})})$ w.r.t.  $\mathbf{e}_i$, \ie $\mathbf{w} = \mathbf{e}_1 + \mathbf{e}_2$ implies $\mathcal{I}(\mathbf{w}) = \mathcal{I}(\mathbf{e}_1)+\mathcal{I}(\mathbf{e}_2)$.

% $\mathcal{I}_\text{eop/dp}(\mathbf{e}_i)<0$ implies a minimize of $f_\text{eop}^\mathcal{V}$ or $f_\text{dp}^\mathcal{V}$, and accordingly, $\mathcal{I}_\text{util}(\mathbf{e}_i)\leq 0$ implies a non-increasing of the utility function $\sum_{z_j}\nabla_\theta\ell(z_j;\hat{\theta(\mathbf{1})})$.

We study the correlation between $\mathcal{I}_\text{eop/dp}$ and $\mathcal{I}_\text{util}$ over the reweighing of $\mathcal{T}$. We first state our assumption for the remaining gradient of $z_i$ at $\opt$.
\begin{restatable}{assumption}{rank}
\label{full}
The gradient matrix of training samples in $\mathcal{T}$ at $\hat{\theta}(\mathbf{1})$:
$\begin{bmatrix}
\nabla\ell_\theta(z_1;\opt) \\
\vdots \\
\nabla\ell_\theta(z_{N_{\mathcal{T}}};\opt) \\
\end{bmatrix}\in\RR^{N_\mathcal{T} \times D}$ has rank $D$.
\end{restatable}
\cref{full} is easy to satisfy if $N_\mathcal{T}\gg D$ in a general training case, \ie one set a model with a proper dimension to fit an adequate number of training samples. The assumption is mild for large-scale data with sufficient variety.

\begin{restatable}{theorem}{eop}{}\label{thm:eop}
If $f_\text{eop}^\mathcal{V}(\hat{\theta}(\mathbf{1}))$ is not in local optimum, $\nabla_\theta f_\text{eop}^\mathcal{V}(\opt)$ and $\sum_{j}\nabla_\theta \ell(z_j;\hat{\theta}(\mathbf{1}))$ are linearly independent, then under assumptions of influence function and \cref{full}, there are conical combinations $\mathbf{w}$ of $\{\mathbf{e}_i\}_{i=1}^{N_\mathcal{T}}$ to construct a reweighing for training samples such that
$\mathcal{I}_\text{eop}(\mathbf{w})<0$ and $\mathcal{I}_\text{util}(\mathbf{w})\leq 0$, with ${\|\mathbf{w}\|}_\infty\leq 1$.
\end{restatable}
\begin{remark}
\cref{thm:eop} states that, if the loss of Equal Opportunity on $\mathcal{V}$ remains a space to improve, then there exists reweighing on $\mathcal{T}$ to improve $f_\text{eop}^\mathcal{V}$ but will not increase the utility loss under some conditions. $\mathbf{w}$ is an element-wise non-negative vector and have an upper bound on entries, which is a nice property we will use in our algorithmic design in the next section. Proofs can be found in \cref{sec:proof}.
\end{remark}
Here we see a proper downweighting can be expected to enhance fairness while keeping the utility not going worse at the influence function level. This finding raises an opportunity to \textit{achieve fairness at no utility cost}. Similarly, we present a corollary that states the same conclusion for Demographic Parity.
\begin{restatable}{corollary}{cordp}{}\label{cor:dp}
If $\nabla_\theta f_\text{dp}^\mathcal{V}({\opt})\neq 0$, $\nabla_\theta f_\text{dp}^\mathcal{V}({\opt})$ and $\sum_{j}\nabla_\theta \ell(z_j;\hat{\theta}(\mathbf{1}))$ are linearly independent, then under assumptions of influence function and \cref{full} satisfied, there are conical combinations $\mathbf{w}$ of $\{\mathbf{e}_i\}_{i=1}^{N_\mathcal{T}}$ to construct a reweighing for training samples such that $\mathcal{I}_\text{dp}(\mathbf{w})<0$ and $\mathcal{I}_\text{util}(\mathbf{w})\leq 0$, with ${\|\mathbf{w}\|}_\infty\leq 1$.
\end{restatable}
Note that a classifier that satisfies Equal Opportunity with an optimal predictive utility performance, might not be optimal for Demographic Parity. A perfect classifier will still encounter a tradeoff between Demographic Parity and predictive utility induced by the difference in base rate between sensitive groups. \cref{cor:dp} is not against this impossibility since it only declares a cost-free improvement on $f_\text{dp}^\mathcal{V}$ when it is not locally optimal.

\textbf{Discussion on Fairness-Utility Tradeoff\ } There is literature theoretically discussing the intrinsic tradeoff between fairness and utility, and such a tradeoff has been empirically revealed in many experiments. \citet{zhao2019inherent} point out the tradeoff between Demographic Parity and the joint error rate across sensitive groups, and~\citet{menon2018cost} also characterize the accuracy-fairness tradeoff holding for an arbitrary classifier. However, they discuss under the context of a given fixed distribution that does not correspond with us. Our findings, from a reweighing perspective, match with the conclusion in~\citet{dutta2020there} that characterizing with Chernoff information from information theory, there exists an ideal distribution where fairness and utility are in accord. However, they do not consider a more practical model learning and inference setting, but primarily restrict to likelihood ratio detectors with synthetic experiments. Our data reweighing via influence function lets us move beyond a bias distribution and release the tradeoff. 

%-------------------------------------------------------------------------

\section{Fairness via Data Reweighing}
\label{sec:reweigh}

The foregoing theoretical analysis raises an opportunity to mitigate the unfairness towards different notions while preserving the predictive utility. In this section, we concretize this idea and convert it into an algorithm to find the optimal weights $\mathbf{w}^*$ through linear programming.

Having $\mathcal{I}_\text{fair}(\mathbf{e}_i)$ and $\mathcal{I}_\text{util}(\mathbf{e}_i)$ in hands, the ideal reweighing is to completely close the gap towards a fairness notion, \eg Equal Opportunity, or reach a user-defined threshold, while preserving the utility not decrease. The searching for $\mathbf{w}^*$ can be cast into a linear program as follows:
\begin{mini}
{}{\sum_{i}\ w_i}{}{}
\addConstraint{\sum_{i}}{\ w_i\mathcal{I}_\text{fair}(\mathbf{e}_i)=-f_\text{fair}^{\mathcal{V}}}{}{}
\addConstraint{\sum_{i}}{\ w_i\mathcal{I}_\text{util}(\mathbf{e}_i)\leq 0}{}{}
\addConstraint{}{w_{i}}{\in[0,1].}{}
\label{lin:1}
\end{mini}
\cref{lin:1} tends to minimize the total amount of perturbation from an initially uniform weight. This prevents downweighting too many samples, thus keeping the generalization capacity from the validation set to the test set. The first subjection is to find the weights that perfectly close the fairness gap in~\cref{eq:fair}, where the minimum of an absolute value is zero or it can also be set to a user-defined threshold. The second subjection is to keep the utility loss non-increasing. In the last subjection we set a range for $w_i$ from 0 to 1, inheriting the property of $\mathbf{w}$ in~\cref{thm:eop}. We set the lower bound as 0 to ensure all the reweighing is downweighting instead of upweighting. Here an upweighting is definitely practicable which emphasizes some particular samples and could achieve the same effect as downweighting on other samples. For the simultaneous upweighting and downweighting, it leads the objective function to be the sum of a set of absolute values between $w_i$ and 1. The linear programming problem with absolute values needs to be transformed into a standard one by introducing additional variables~\cite{dantzig2016linear}. Also, we do not observe a conspicuous benefit from introducing upweighting in practice. For simplicity and efficiency, we only consider the downweighting for data reweighing. The upper bound of weights prevents the linear programming from turning a sample into a negative sample which affords negative loss. This is for training stability concerns and avoids the reweighing overly concentrating on one sample.

However, \cref{lin:1} may not offer a feasible solution since there might exist a case that $\min_{\mathbf{w}}\sum_{i}{\ w_i\mathcal{I}_\text{fair}(\mathbf{e}_i)}>-f_\text{fair}^{\mathcal{V}}$ with the rest constraints held. In this case, we substitute \cref{lin:1} with \cref{lin:2} stated as follow:
\begin{mini}
{}{\sum_i\ w_i\mathcal{I}_\text{fair}(\mathbf{e}_i)}{}{}
\addConstraint{\sum_i}{\ w_i\mathcal{I}_\text{util}(\mathbf{e}_i)\leq 0}{}{}
\addConstraint{\sum_i}{\ w_i\leq\alpha N_\mathcal{T}}{}{}
\addConstraint{}{w_{i}}{\in[0,1],}{}
\label{lin:2}
\end{mini}
with $\alpha\in(0,1]$. \cref{lin:2} rotates the first constraint to its objective function and replenishes a constraint on the quantity of perturbation w.r.t the total number of training samples. $\alpha$ here is a hyperparameter indicating the proportion of weights to be changed. Note that we set a higher priority of \cref{lin:1} over \cref{lin:2}, since if $\min_{\mathbf{w}}\sum_{i}{\ w_i\mathcal{I}_\text{fair}(\mathbf{e}_i)}<-f_\text{fair}^{\mathcal{V}}$, the objective values goes to negative and cause unfairness again, \textit{i.e.,} it flips the original unprivileged group into a privileged group. If one wants to support \cref{lin:2} as the priority, a lower bound of $-f_\text{fair}^\mathcal{V}$ should be added to the objective function but that can cause dual degeneracy for a minimization problem in LP.

In practice, although the solution seems to be optimal in terms of influential approximation, $\mathbf{w}^*$ cannot imply absolute fairness at no utility cost. The error comes from two sides. (1) Even though influence function is almost precise for the removal of an individual sample, transferring the influence function to the actual influence still suffers some additional errors related to the quantity of perturbation. \citet{koh2019accuracy} empirically reveal that when removing a group of samples from the training set, influence function tends to underestimate or overestimate the actual influence, but still keeps a high correlation. Colloquially, the larger the group removed from the training set, the more imprecision can be expected. (2) There is a gap between the surrogate function (for differentiable purpose) and the real value of Equal Opportunity gap or Demographic Parity gap, as well as the utility. As a remedy, we complement some extra hyperparameters in the constraints of \cref{lin:1} to alleviate such errors, and we restate it as follows:
\begin{mini}
{}{\sum_{i} w_i}{}{}
\addConstraint{\sum_{i}}{w_i\mathcal{I}_\text{fair}(\mathbf{e}_i)\leq-(1-\beta)\ell_\text{fair}^{\mathcal{V}}}{}{}
\addConstraint{\sum_{i}}{w_i\mathcal{I}_\text{util}(\mathbf{e}_i)\leq\gamma(\min_\mathbf{v}\sum_{i}v_i\mathcal{I}_\text{util}(\mathbf{e}_i))}{}{}
\addConstraint{ }{w_{i}}{\in[0,1].}{}
\label{lin:relax}
\end{mini}
We linearly tighten or relax the constraint with hyperparameters $\beta$ and $\gamma$ since influence function under group effect could enlarge the scale of deflection but still under a high correlation. If there is a deflection, we can observe it on the validation set and regulate the objectives to compensate for the group effect. We conduct a grid search for $\beta$ and $\gamma$ on the validation set, and demonstrate the performance on the test set. We summarize the algorithmic pipeline in \cref{alg:reweigh} and show its property of feasibility.
\begin{restatable}{corollary}{alg}{}\label{cor:feasible}
If assumptions in \cref{thm:eop} and \cref{cor:dp} are satisfied, then \cref{alg:reweigh} could reach a feasible and non-trivial solution.
\end{restatable}
\begin{remark}
A trivial solution means the solved $\mathbf{w}=0$ that does not reweigh any training samples.
\end{remark}

\begin{algorithm}[tb]
   \caption{No Utility-Cost Fairness via Data Reweighing}
   \label{alg:reweigh}
\begin{algorithmic}[1]
   \STATE {\bfseries Input:} Training set $\mathcal{T}=\{z_i\}_{i=1}^{N_\mathcal{T}}$ , validation set $\mathcal{V}=\{z_j,a_j\}_{j=1}^{N_\mathcal{V}}$.
   \STATE Train $\hat{\theta}(\mathbf{1}) = \argmin_{\theta}\sum_{i} \ell(z_i;\theta)$ by \cref{eq:train};
   \STATE Compute $f_\text{fair}^\mathcal{V}$ by \cref{eq:fair};
   \STATE Compute $\sum_j\nabla_\theta\ell(z_j;\hat{\theta}(\mathbf{1}))$;
   \STATE Compute Hessian vector product $\mathbf{H}_{\hat{\theta}(\mathbf{1})}^{-1}\nabla_\theta\ell(z_i;\hat{\theta}(\mathbf{1}))$ for every training sample $i\in\mathcal{T}$;
   \STATE Compute $\mathcal{I}_\text{fair}(\mathbf{e}_i)$ and $\mathcal{I}_\text{util}(\mathbf{e}_i)$ for every training sample $i\in\mathcal{T}$ by \cref{eq:infl_util,eq:infl_fair};
   \STATE Solve the linear programming problem in \cref{lin:relax};
   \IF{\cref{lin:relax} is infeasible}
   \STATE Solve the linear programming problem in \cref{lin:2};
   \ENDIF
   \STATE Retrain the model with $\mathbf{w}^*$ by \cref{eq:retrain} and obtain $\hat{\theta}(\mathbf{1}-\mathbf{w}^*)$;
   \STATE Evaluate the test set with $\hat{\theta}(\mathbf{1}-\mathbf{w}^*)$.
\end{algorithmic}
\end{algorithm}

%-------------------------------------------------------------------------

\section{Related Work}
\label{sec:relate}

We chronologically review the pre-processing methods for algorithmic fairness, and introduce related works on influence function. We include many of these pre-processing methods in experimental comparison.

\textbf{Fair Algorithms with Pre-processing\ } Previous works develop pre-processing techniques to ensure a nice property that the fairness-oriented improvement is independent of existing learning pipelines. \citet{kamiran2012data} propose several reweighing and label flipping approaches based on data statistics to expand the underrepresented group or benefit the unprivileged group. Some of these approaches further consider the original predictive confidence. \citet{zemel2013learning} learn a fair intermediate representation for data towards both group and individual fairness. \citet{feldman2015certifying} satisfy Demographic Parity by transforming the input data into fair features while preserving their original rank. \citet{calmon2017optimized} follow the data fair transformation and optimize the transformation with individual least distance constraints. \citet{krasanakis2018adaptive} and \citet{jiang2020identifying} introduce reweighing approaches and update weights iteratively through a continuous learning process. The iterative update asks for retraining the model within each learning iteration. \citet{wang2019repairing} use a descent algorithm to learn a counterfactual distribution from data to close the fairness gap for a black-box predictor, and build the pre-processing via optimal transport. \citet{yan2020fair} and \citet{lahoti2020fairness} focus on a case where the sensitive attribute is missing. \citet{yan2020fair} use clustering to find out underrepresented groups and complements them with nearest neighbor searching. \citet{lahoti2020fairness} considers Rawlsian Max-Min Fairness with a special interest in the worst-case accuracy. Many works consider fairness subjecting to predictive utility constraints in design, and the intention to preserve the utility helps to popularize fair algorithms. Our algorithmic design considers both fairness and predictive utility via a one-pass reweighing approach that is friendly for subsequent learning pipelines which might be time-consuming. We granularly characterize each training sample using their fairness and utility influence, and compute individual weights by solving linear programs.

\textbf{Influence Function\ } Influence function originates from diagnostic in statistics~\cite{cook1977detection,cook1980characterizations}. It approximates the actual effect brought by the removal of training points from models or other perturbations on data. Influence function in machine learning has been used to study model robustness~\cite{christmann2004robustness,hampel2011robust,liu2014efficient}. Recently, \citet{koh2017understanding} extend influence function to various large-scale machine learning models, and introduce its applications in adversarial attack, data interpretation, and label fixing. More works follow up to develop this statistical tool. ~\citet{giordano2019swiss} provides finite-sample error bounds on the leave-k-out case of the asymptotic results. \citet{koh2019accuracy} characterize the error when removing a group of data and provide interesting empirical findings. They reveal that when influence function is measuring the effect of removing a random or a certain group of samples, the absolute and relative error will be large, but still correlates well with the actual influence. In this work, we consider an innovative application of influence function in considering both algorithmic fairness and utility.

%-------------------------------------------------------------------------

\section{Experiment}
\label{sec:exp}

%-------------------------------------------------------------------------

\subsection{Dataset}

We use the following real-world tabular datasets for experiments~\cite{Dua:2019}. We provide statistics in~\cref{sec:data}. \textbf{Adult}. The Adult dataset~\cite{adult} contains 45,222 census personal records. It includes 14 attributes such as age, education, race, etc. The goal is to predict if the personal annual income exceeds 50k. We set gender as the sensitive attribute. \textbf{Compas}. The Compas dataset~\cite{compas} records information like criminal history, jail and prison time, demographic, etc. The dataset is to predict a recidivism risk score for defendants. We set race as the sensitive attribute. \textbf{Communities and Crime}. The Communities and Crime dataset~\cite{redmond2002data} describes communities with the percent of the population considered urban, the median family income, etc. The goal is to predict violent crimes and we set the percentage of the black population as the sensitive attribute. \textbf{German Credit}. The German Credit dataset~\cite{german} classifies people as good or bad credit risks using their profile and history. We set age as the sensitive attribute with a threshold at 30.

%-------------------------------------------------------------------------

\subsection{Implementation and Protocol}

For fair classification, we consider pre-processing baselines: Massaging, Reweigh, Preferential from~\citet{kamiran2012data}, and Disparate Impact Remover (Dis. Remover)~\cite{feldman2015certifying}, Label Bias~\cite{jiang2020identifying}, and Adversarial Reweighted Learning (ARL)~\cite{lahoti2020fairness}. We also involve two in-processing adversarial training methods: Adversarial Debiasing (Adv.)~\cite{zhang2018mitigating} and Conditional Adversarial Debiasing (Cond. Adv.)~\cite{zhao2019conditional}. For comprehensiveness, we consider two kinds of base models: Logistic Regression and two-layer non-linear Neural Networks, where we use Logistic Regression for all four datasets and use Neural Networks for the Adult and Compas datasets due to their large sample sizes. Most pre-processing approaches work independently of the base model and can be wrapped on both Logistic Regression and Neural Networks, while adversarial training methods only work for Neural Networks. We equip our methods to Neural Networks by only computing the influence to the last layer of Neural Networks and retraining this part of weights, thus the convexity is guaranteed. The specific parameters of base models and input data are used exactly the same across all baselines and our methods. Input data are standardized by removing the mean and scaling to unit variance. For methods with a hyperparameter to directly control the fairness-utility tradeoff, we tune the hyperparameter and show the Pareto curve in figures. Linear programs in~\cref{alg:reweigh} are solved using Gurobi~\cite{gurobi} under an academic license. We divide all the datasets into training set (60\%), validation set (20\%), and test set (20\%), except for the Adult dataset that has a pre-defined split on training/validation/test set. More details are reported in~\cref{sec:hyper}.

We explicitly compute the Hessian matrix and its inverse to get the influence function. A typical complexity for computing Hessian is $\mathcal{O}(nd^2)$, where $n$ and $d$ are the numbers of samples and model dimension, and its inversion takes $\mathcal{O}(d^3)$. To address high-dimensional models, we can apply conjugate gradients or stochastic estimation of Hessian-vector products, resulting in $\mathcal{O}(nd)$~\cite{koh2017understanding}.

%-------------------------------------------------------------------------

\begin{figure*}
\centering
  \includegraphics[width=1.79\columnwidth]{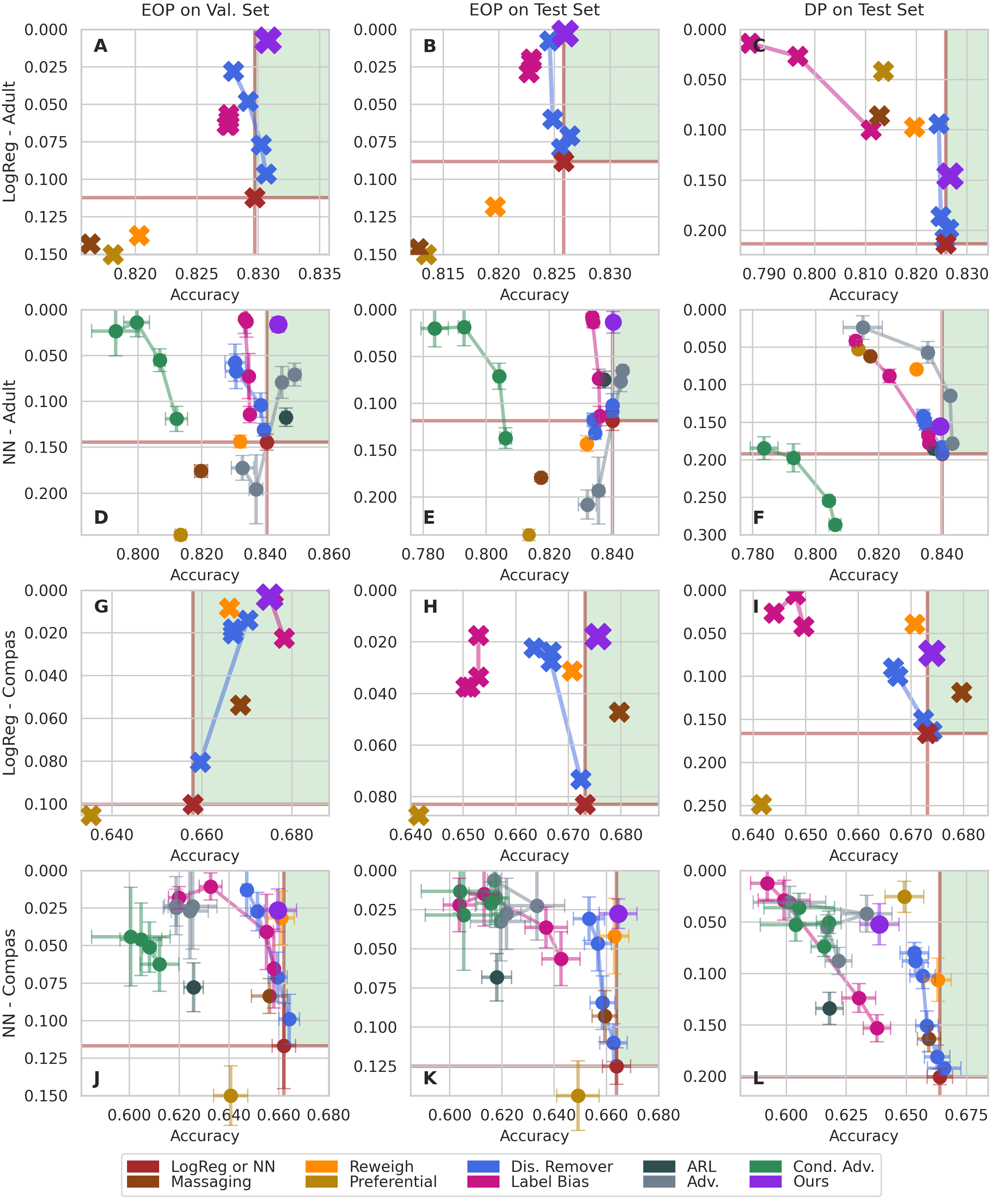}
  \caption{Experimental results of five pre-processing and two in-processing fair classification methods on the Adult and Compas datasets. For each figure, we indicate the corresponding base model, experimental dataset, fairness metric, and evaluation set in the left and top header (EOP$\rightarrow$ Equal Opportunity, DP$\rightarrow$ Demographic Parity). Y-axis for fairness is inverted and shows the absolute value of the gap in fairness between the privileged and unprivileged groups (the higher position means the smaller fairness metric value and the better performance towards fairness), while X-axis shows the predictive accuracy. According to the utility and fairness performance of the base model LogReg (Logistic Regression) or NN (Neural Networks), we plot a horizontal and a vertical line in each figure and divide the space by fairness and utility results into four regions, where the space in green means a fairer and more accurate model compared to the base model. A point closer to the top right indicates better performance in both utility and fairness. The values of some points by the Preferential method are out of the current scale of these figures; for better visualization, we put these points at the boundary of a figure. We do not observe randomness for the Logistic Regression model, while we plot the standard deviation for Neural Networks with results obtained by five random seeds. For methods with a hyperparameter to control the tradeoff between fairness and utility, the line connecting the same method indicates the monotonously increased change on the hyperparameter. For details please refer to~\cref{sec:hyper}.}
  \label{fig:main}
\end{figure*}

\begin{figure*}
\centering
  \includegraphics[width=1.85\columnwidth]{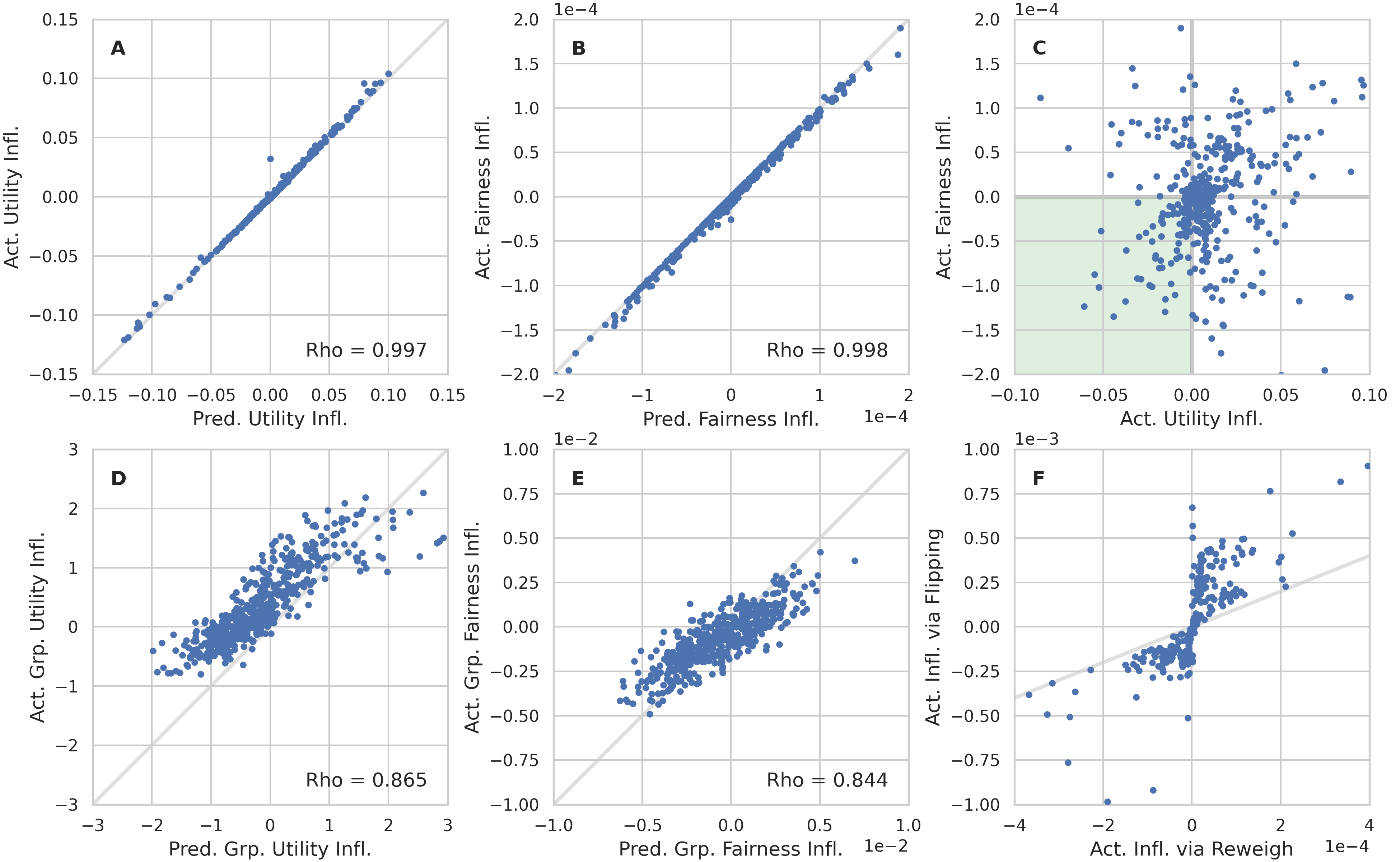}
  \vspace{-4mm}
  \caption{Studies on influence function and its actual effect. All results are obtained on the Adult dataset. \textbf{A} and \textbf{B}: Predictive influence v.s. actual influence by leave-one-out retraining in terms of utility and Equal Opportunity loss; \textbf{C}: Actual influence of utility loss v.s. Equal Opportunity loss by leave-one-out retraining. \textbf{A}, \textbf{B} and \textbf{C} are conducted on 500 randomly selected training samples; \textbf{D} and \textbf{E}: Influence function v.s. actual effect by leave-group-out retraining in terms of utility and Equal Opportunity loss with a group size equal to 250. \textbf{F}: Actual influence of Equal Opportunity loss by target label flip and retraining v.s. reweighing and retraining from our algorithm. `Rho' indicates Pearson correlation coefficient in figures.}
  \label{fig:infl}
  \vspace{-4mm}
\end{figure*}

%-------------------------------------------------------------------------

\subsection{Experimental Comparison}

We present our experimental results on the Adult and Compas datasets in~\cref{fig:main}, and defer other results to~\cref{sec:extra}. Through experiments we have several findings:  (1) In general, except in Figure \textbf{G} where the majority of these fair classifiers deliver simultaneous and non-trivial improvements in both fairness and utility, the experimental results of baseline methods are mainly located at the top left region, indicating that their improved fairness is achieved at a non-negligible cost of utility. (2) Specifically, heuristic pre-processing methods including Massaging, Reweigh, and Preferential fail to guarantee a stable improvement over fairness, and sometimes even decrease Equal Opportunity. The learning approaches Disparate Impact Remover and Label Bias induce a tradeoff between fairness and utility, and unfortunately lead to either unsatisfied fairness or an unacceptable utility cost in many cases. The adversarial approaches Adversarial Debiasing and Conditional Adversarial Debiasing also bring an obligatory tradeoff so that they need careful tuning. Moreover, they induce a large variance in Neural Networks model. (3) In most cases (See \textbf{A}-\textbf{I}), our method is able to improve fairness while keeping the utility unchanged or even slightly improved. This nice property and its pre-processing fashion can play a significant advantage in popularizing fair algorithms. These empirical results also verify our theoretical findings. It is worthy to note that our method cannot escape the upper bound of Demographic Parity at no utility cost since this scenario resists the impossibility from the difference in the base rate. In a few cases (See \textbf{J}-\textbf{K}), our method sacrifices a little utility to greatly improve the fairness. Due to the nature of influence approximation, when removing a group of samples from the training set, there exists a gap between actual and predicted influence. There is still some deviation even though we add hyperparameters to fulfill the gap in~\cref{lin:relax}.

%-------------------------------------------------------------------------

\subsection{Sample Influence and Actual Effect}

We investigate the influence function and its actual influence in~\cref{fig:infl}. In \textbf{A} and \textbf{B}, we visualize influence function and corresponding actual change after leave-one-out retraining, and show that influence function can predict the actual influence with high precision for both utility and fairness. In \textbf{C}, we show the distribution of actual influence for individual samples. Some parts of points are located at the green regions, meaning that removing or downweighting these points can simultaneously improve fairness and utility. This is well in accord with our~\cref{thm:eop}. In \textbf{D} and \textbf{E}, we show how the influence function differs from removing an individual sample to removing a group of samples. The influence function still remains a high correlation but results in a larger error. The imprecision justifies our design in~\cref{lin:relax} where we use hyperparameters to mitigate this effect. In \textbf{F}, we compare the influence of removing samples to target label flipping, complementing the label bias hypothesis in the introduction. Flipping the label can incur a larger change in loss value compared to simply reweighing them.

%-------------------------------------------------------------------------

\section{Conclusion}
\label{sec:con}

We proposed a data reweighing approach for improving algorithmic fairness and utility. We granularly measured the influence of every training sample towards fairness and utility on a validation set. We demonstrated that under some mild assumptions, there exists some reweighing on training samples that can improve fairness at no utility cost. We solved the reweighing strategy through linear programming with constraints of both fairness and utility, and guaranteed its feasibility. We empirically verified our algorithms on multiple tabular datasets and showed that the traditional fairness-utility tradeoff could be released in most cases.

%-------------------------------------------------------------------------

\bibliography{ref}
\bibliographystyle{icml2022}

%-------------------------------------------------------------------------

%%%%%%%%%%%%%%%%%%%%%%%%%%%%%%%%%%%%%%%%%%%%%%%%%%%%%%%%%%%%%%%%%%%%%%%%%%%%%%%
%%%%%%%%%%%%%%%%%%%%%%%%%%%%%%%%%%%%%%%%%%%%%%%%%%%%%%%%%%%%%%%%%%%%%%%%%%%%%%%
% APPENDIX
%%%%%%%%%%%%%%%%%%%%%%%%%%%%%%%%%%%%%%%%%%%%%%%%%%%%%%%%%%%%%%%%%%%%%%%%%%%%%%%
%%%%%%%%%%%%%%%%%%%%%%%%%%%%%%%%%%%%%%%%%%%%%%%%%%%%%%%%%%%%%%%%%%%%%%%%%%%%%%%
\newpage
\appendix
\onecolumn

%-------------------------------------------------------------------------

\section{Proof and Discussion}
\label{sec:proof}
\rank*

We discuss the feasibility of the assumption here. The assumption requires that we have training samples such that the gradient vector matrix has full rank. This is a mild assumption for a large training dataset with sufficient sample diversity and a proper model dimension. The assumption could be violated when (1). the model has a very large and improper dimension such that some parameters are not even activated by the training data, and (2). there are very limited training points. For the second scenario, consider an extreme case if we only have one training sample for the model, therefore we cannot reweight this sample to change the model's parameter, but luckily it is not likely to happen in general cases. In Figure $\mathbf{C}$ from~\cref{fig:infl} we show some hard removals of samples are bringing positive impact to both utility and fairness, which confirm our theorem and the feasibility of this assumption.

\eop*
\begin{proof}

For completeness, we restate the Equal Opportunity loss over the validation set $\mathcal{V}$ in \cref{eq:fair} as follows:
\begin{equation}
\begin{aligned}
    f_\text{eop}^\mathcal{V}(\opt)&=|\Exp_\mathcal{V}[\ell(z;\opt)\ |\ a=1,y=1] - \Exp_\mathcal{V}[\ell(z;\opt)\ |\ a=0,y=1]| \\\
    &=|\frac{1}{|\mathcal{V}_{a=1,y=1}|} \sum_{j:a_j=1,y_j=1}\ell(z_j;\hat{\theta}(\mathbf{1})) - \frac{1}{|\mathcal{V}_{a=0,y=1}|} \sum_{j:a_j=0,y_j=1}\ell(z_j;\hat{\theta}(\mathbf{1}))|>0,
\end{aligned}
\end{equation}
where $\mathcal{V}_{a=1}$ denotes the set conditional on $a=1$, and $|\mathcal{V}|$ is the cardinality of set $\mathcal{V}$. We have $\nabla_\theta f_\text{eop}^\mathcal{V}(\opt)\neq 0$ since $\theta f_\text{eop}^\mathcal{V}(\opt)$ is not in local optimum, and would guarantees $\nabla_\theta {f_\text{eop}^\mathcal{V}(\opt)}^\top\hess\neq 0$ in $\mathcal{I}_\text{eop}$.

Now let $\mathbf{G}=\begin{bmatrix}
\nabla\ell_\theta(z_1;\opt) \\
\vdots \\
\nabla\ell_\theta(z_{N_{\mathcal{T}}};\opt) \\
\end{bmatrix}\in\RR^{N_\mathcal{T} \times D}$ with full rank $D$ (\cref{full}), we can write the influence function towards Equal Opportunity and utility in \cref{eq:infl_fair} and \cref{eq:infl_util} as follows:

\begin{equation}
\begin{aligned}
    \mathcal{I}_\text{eop}(\mathbf{w})&=(\nabla_\theta {f_\text{eop}^\mathcal{V}(\opt)}^\top\hess)({\mathbf{G}}^\top\mathbf{w}), \\
    \mathcal{I}_\text{util}(\mathbf{w})&=(\sum_j\nabla_\theta\ell(z_j;\opt)^\top\hess)({\mathbf{G}}^\top\mathbf{w}).
\end{aligned}
\end{equation}

Since $\nabla_\theta {f_\text{eop}^\mathcal{V}(\opt)}$ and $\sum_j\nabla_\theta\ell(z_j;\opt)$ are linearly independent, we have $\nabla_\theta {f_\text{eop}^\mathcal{V}(\opt)}^\top\hess$ and $\sum_j\nabla_\theta\ell(z_j;\opt)^\top\hess$ to be linearly independent, since $\hess$ is non-singular. Because $\mathbf{G}$ has rank $D$, so the gradient vectors in $\mathbf{G}$ span $\RR^{D}$, thus we can always find $\mathbf{w}$ on $N_\mathcal{T}$ training samples such that $\mathcal{I}_\text{eop}(\mathbf{w})<0$ and $\mathcal{I}_\text{util}(\mathbf{w})\leq 0$.

Recall the first-order optimality $\sum_i\nabla\ell(z_i;\opt)=0\rightarrow\mathbf{1}^\top\mathbf{G}=0$ for the model. Once we have $\mathbf{w}$ satisfy $\mathcal{I}_\text{eop}(\mathbf{w})<0$ and $\mathcal{I}_\text{util}(\mathbf{w})\leq0$ on $N_{\mathcal{T}}$, we should have
\begin{equation}
\begin{aligned}
    \mathcal{I}_\text{eop}(\mathbf{w})&=(\nabla_\theta {f_\text{eop}^\mathcal{V}(\opt)}^\top\hess)(\mathbf{G}^\top(\mathbf{w}+c\mathbf{1}))<0, \\
    \mathcal{I}_\text{util}(\mathbf{w})&=(\sum_j\nabla_\theta\ell(z_j;\opt)^\top\hess)(\mathbf{G}^\top(\mathbf{w}+c\mathbf{1}))\leq 0,
\end{aligned}
\end{equation}
where $c$ is an arbitrary real number. If $\mathbf{w}$ has negative entries, let $c=-\min_{i}{w_i}$  and $\mathbf{w}\leftarrow(\mathbf{w} + c\mathbf{1})$ helps to convert all entries into non-negative. A normalization $\mathbf{w}\leftarrow{\mathbf{w}}/{\|\mathbf{w}\|}$ helps to bound $\|\mathbf{w}\|_\infty\leq 1$, hence completing the proof.
\end{proof}

\cordp*
\begin{proof}
The proof follows the proof for~\cref{thm:eop} by replacing $\nabla_\theta {f_\text{eop}^\mathcal{V}(\opt)}$ with $\nabla_\theta {f_\text{dp}^\mathcal{V}(\opt)}$.
\end{proof}
% \begin{proof}
% With out loss of generality, we assume $a=1$ is the privileged group that enjoys a higher averaged predictive rate $\hat{y}$ on $\mathcal{V}$. Following the chain rule we have
% \begin{equation}
% \begin{aligned}
%     \nabla_\theta f_\text{dp}^\mathcal{V}(\opt)&=\frac{1}{|\mathcal{V}_{a=1}|}\sum_{j:a_j=1}\frac{\partial\hat{y}_j}{\partial\opt} - \frac{1}{|\mathcal{V}_{a=0}|}\sum_{j:a_j=0}\frac{\partial\hat{y}_j}{\partial\opt}, \\
%     \sum_j\nabla_\theta\ell(z_j;\opt)&=\sum_{j:a_j=1}\frac{\partial\ell(z_j;\opt)}{\partial\hat{y}_j}\frac{\partial \hat{y}_j}{\partial\opt} + \sum_{j:a_j=0}\frac{\partial\ell(z_j;\opt)}{\partial\hat{y}_j}\frac{\partial \hat{y}_j}{\partial\opt}.
% \end{aligned}
% \end{equation}
% The linear independence of $\sum_{j:a_j=1}\partial\hat{y}_j/\partial\opt$ and $\sum_{j:a_j=0}\partial\hat{y}_j/\partial\opt$ induces the linear independence of $\nabla_\theta f_\text{dp}^\mathcal{V}(\opt)$ and $\sum_j\nabla_\theta\ell(z_j;\opt)$. The rest follows the proof of \cref{thm:eop}.
% \end{proof}

\alg*
\begin{proof}
For any $\alpha>0$, following the results from \cref{thm:eop} and \cref{cor:dp}, we shall always find a non-negative $\mathbf{w}^*$ satisfying constraints in~\cref{lin:2} and have $\mathcal{I}_\text{fair}(\mathbf{w}^*)<0$. This induces the objective function value in~\cref{lin:2} with $\mathbf{w}^*$ smaller than 0. Note that a trivial solution with an all-zero $\mathbf{w}$ makes the objective function zero.
\end{proof}

%-------------------------------------------------------------------------

\section{Dataset Statistics}\label{sec:data}

We summarize some key statistics of four datasets we use in experiments in~\cref{tab:data}. For two numerical sensitive variables` \%Black Popluation' and `Age,' we use them and set a threshold to divide the privileged and unprivileged groups. `Group Pos. Rate' calculates the proportion of samples which has a positive label in the privileged and unprivileged group, respectively. `$\ell_2$ reg.' is L2 regularization strength for the Logistic Regression model, obtained by a grid search over the validation set. The former number is L2 regularization normalized by the number of training samples, while the latter one is the value we set for models. We set 1e-3 as the L2 regularization for Neural Networks as default.

\begin{table}[h]
\vspace{-6mm}
\caption{Dataset Statistics}
\vspace{1mm}
\label{tab:data}
\begin{center}
\begin{small}
\begin{tabular}{lccccccc}
\toprule
Dataset & \#Sample (Train / Val. / Test) & \#Dim. & Sensitive Attribute & Group Pos. Rate & $\ell_2$ reg. for LogReg \\
\midrule
Adult               & 22,622 / 7,540 / 15,060 & 102 & Gender - Male / Female & 0.312 / 0.113 & 1.00e-4 $\rightarrow$ 2.26 \\
Compas              & 3,700 / 1,234 / 1,233   & 433 & Race - White / Non-white   & 0.609 / 0.518 & 1.00e-2 $\rightarrow$ 37.00 \\
Comm. and Crime    & 1,196 / 399 / 399        & 122 & \%Black Popluation - 0.06 & 0.887 / 0.537 & 2.15e-2 $\rightarrow$ 25.79 \\
German Credit       & 600 / 200 / 200         & 56  & Age - 30           & 0.742 / 0.643 & 9.75e-3$\rightarrow$ 5.85 \\
\bottomrule
\end{tabular}
\end{small}
\end{center}
\end{table}

%-------------------------------------------------------------------------

\section{Hyperparameter Selection}\label{sec:hyper}

We use grid search on the validation to set hyperparameters in~\cref{lin:2} and~\cref{lin:relax}. The procedure for~\cref{lin:relax} is: first, we set $\beta=0$ and $\gamma=0$, and observe the performance of fairness. After reaching fairness to a desirable level, we tune $\gamma$ to control the utility and try to keep its original utility on the validation set. A finer tuning on $\beta$ is conducted after we fix $\gamma$. The minimum interval of $\beta$ and $\gamma$ between values is 0.1, with range $[0,0.9]$ and $[0,0.4]$, respectively. For $\alpha$ in~\cref{lin:2}, increasing it from 0 with interval 0.01 with a maximum value of $0.15$ can reach our results. The final selections are listed as follows.

Adult: LogReg - EOP: $\beta=0.5$, $\gamma=0.2$; NN - EOP: $\beta=0.5$, $\gamma=0.2$; LogReg - DP: $\beta=0.8$, $\gamma=0.3$; NN - DP: $\alpha=0.02$.

Compas: LogReg - EOP: $\beta=0.2,\gamma=0.1$; NN - EOP: $\beta=0.2,\gamma=0.1$; LogReg - DP: $\beta=0.3,\gamma=0.1$; NN - DP: $\beta=0.3,\gamma=0.1$.

Comm.: LogReg - EOP: $\alpha=0.1$; LogReg - DP: $\alpha=0.1$.

German.: LogReg - EOP: $\beta=0.0,\gamma=0.0$; LogReg - DP: $\beta=0.5,\gamma=0.0$.

For baseline methods with a controllable tradeoff, we use the hyperparamters listed as follows: Dis. remover: $\text{repair level}=\{0.25,0.5,0.75,1.0\}$; Label bias: $\text{learning rate}=\{0.05,0.1,0.5,1.0\}$; Adv. and Cond. Adv.: $\alpha=\{0.1,1.0,5.0,10.0\}$.

\begin{figure*}
\centering
  \includegraphics[width=0.35\columnwidth]{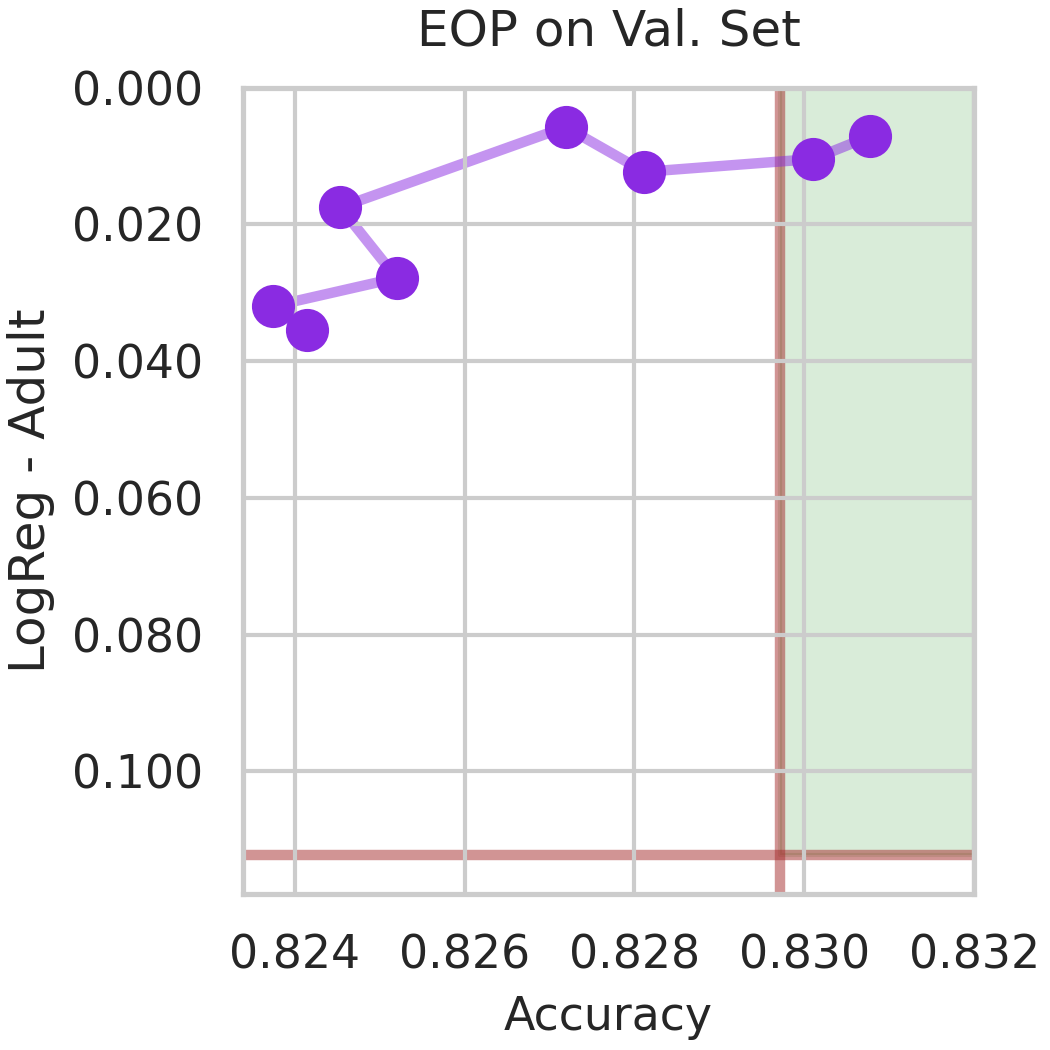}
  \vspace{-4mm}
  \caption{Hyperparameters tuning for~\cref{alg:reweigh}. The lines between nodes show the process of tuning hyperparameters. Initially, the model suffers from group effect of influence function, and the hyperparameters help to mitigate such effect and help the model reach the green region.}
  \label{fig:hyper}
\end{figure*}

We show a hyperparameters tuning process on the Adult dataset with logistic regression in~\cref{fig:hyper}.

%-------------------------------------------------------------------------

\section{Additional Results}\label{sec:extra}

\begin{figure*}
\centering
  \includegraphics[width=0.8\columnwidth]{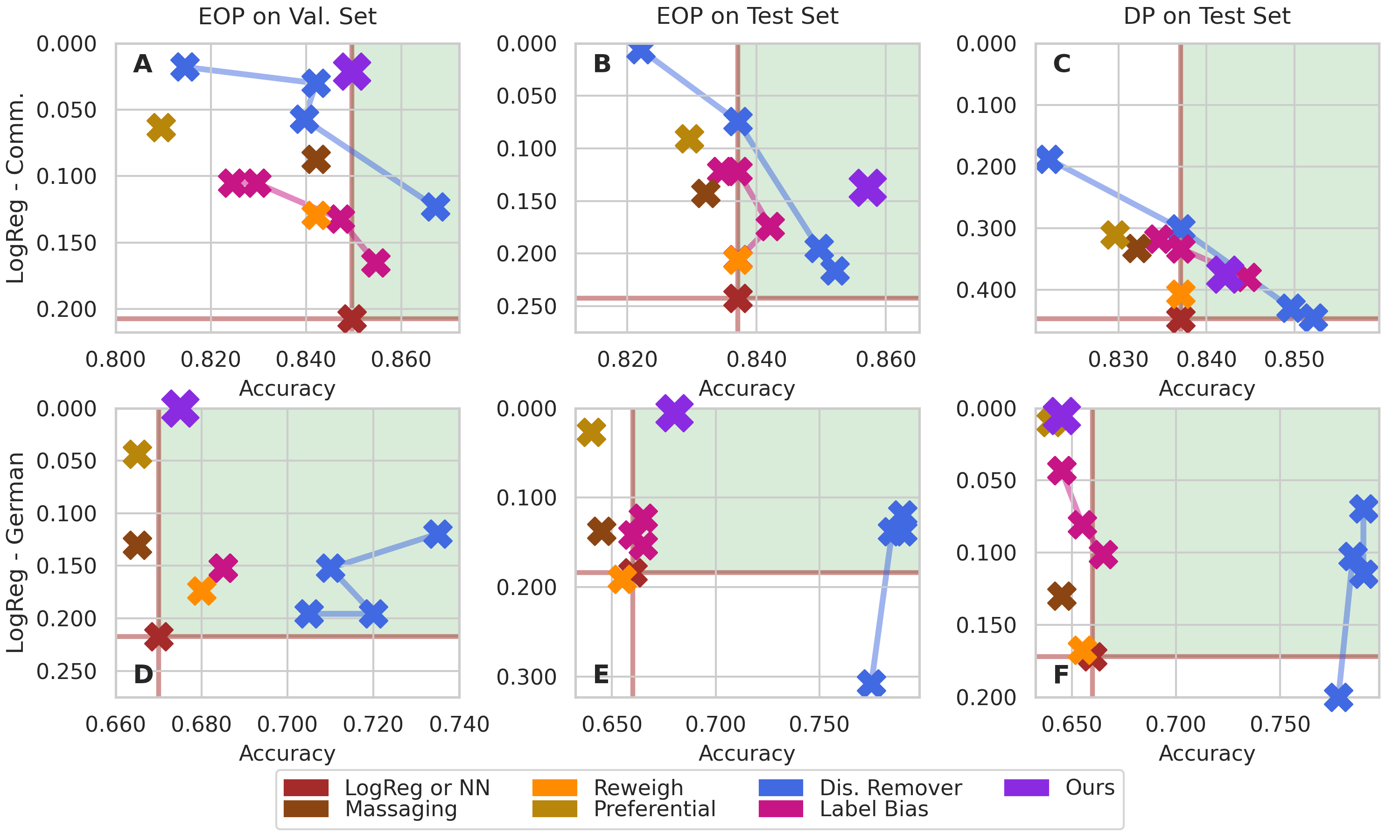}
  \caption{Experimental results on the Communities and Crime and German Credit datasets with LogReg (Logistic Regression) as the base model. For each figure, we indicate the corresponding base model, experimental dataset, fairness metric, and evaluation set in the left and top header (EOP$\rightarrow$ Equal Opportunity, DP$\rightarrow$ Demographic Parity). Y-axis for fairness is inverted and shows the absolute value of the gap in fairness between the privileged and unprivileged groups (the higher position means the smaller fairness metric value and the better performance towards fairness), while X-axis shows the predictive accuracy. According to the utility and fairness performance of the base model, we plot a horizontal and a vertical line in each figure and divide the space by fairness and utility results into four regions, where the space in green means a fairer and more accurate model compared to the base model. A point closer to the top right indicates better performance in both utility and fairness. For methods with a hyperparameter to control the tradeoff between fairness and utility, the line connecting the same method indicates the monotonously increased change on the hyperparameter.}
  \label{fig:extra}
\end{figure*}

We present experimental results for the Community and Crime dataset and the German Credit datasets in~\cref{fig:extra}. Our method works consistently across all datasets, and in most cases, we obtain an improvement in fairness at no utility cost.

\end{document}